\documentclass[12pt]{article}

\makeatletter
\def\set@curr@file#1{%
	\begingroup
	\escapechar\m@ne
	\xdef\@curr@file{\expandafter\string\csname #1\endcsname}%
	\endgroup
}
\def\quote@name#1{"\quote@@name#1\@gobble""}
\def\quote@@name#1"{#1\quote@@name}
\def\unquote@name#1{\quote@@name#1\@gobble"}
\makeatother

\usepackage{latexsym}
\usepackage{verbatim}
\usepackage{graphicx}
\usepackage{mathptmx}
\usepackage{float}
\usepackage{changepage}
\usepackage[normalem]{ulem}
\usepackage{wrapfig}
\usepackage[margin=1.0in]{geometry}
\usepackage{csquotes}
\usepackage{pdfcomment}
\usepackage{caption}
\usepackage{rotating}
\usepackage{times}
\usepackage{hyperref}

\fontfamily{ptm}

\title{\textbf{Training robust anomaly detection using ML-Enhanced simulations} \\
	ASRC Federal Holding Company \\
	}
\author{Philip Feldman, Ph.D.\\
	410.300.7293\\
	philip.feldman@asrcfederal.com
}

\begin{document}
	
\maketitle
\newpage
%\tableofcontents
\newpage

\begin{abstract}
	This paper describes the use of neural networks to enhance simulations for subsequent training of anomaly-detection systems. Simulations can provide edge conditions for anomaly detection which may be sparse or non-existent in real-world data. Simulations suffer, however, by producing data that is \enquote{too clean} resulting in anomaly detection systems that cannot transition from simulated data to actual conditions. Our approach enhances simulations using neural networks trained on real-world data to create outputs that are more realistic and variable than traditional simulations. 
\end{abstract}

\section{\uppercase{Introduction}}
Creating autonomous vehicles that can perform optimally in unusual circumstances is a difficult problem in machine learning (ML). The reason for this is that most data is collected from systems functioning \textit{normally}. For example, it is relatively straightforward to create a training set of typical rush hour traffic by simply equipping cars with cameras and driving them around in cities known to have traffic problems. But this is only a partial solution. In major evacuations, such as those for hurricanes, traffic is often directed to use all available lanes. A self-driving car that is not trained for that possibility can be expected to behave in unpredictable ways. A neural network can easily \enquote{learn} to ignore such corner cases. For example, a network can be trained to drive one mile with perfect (99.9998\%) accuracy if it assumes accidents simply do not happen~\cite{usdot_traffic_2018}.

This also happens with ML-based anomaly detection and response systems. The vast number of vehicles, from automobiles to satellites behave nominally for the vast majority of their functional lifespan. Training for degraded modes requires vast amount of data being collected in a large number of malfunctioning states. Often, this data does not exist in sufficient quantity, and would be expensive to produce. One can imagine the paperwork required to slowly and rigorously destroy a collection of multi-million dollar vehicles simply to train their diagnostic systems.

%These types of errors that result from complex, hidden interactions are known as \textit{normal accidents}~\cite{perrow2011normal}. 

An effective solution to this problem is to use simulations~\cite{pretorius2013simulating, zadok2019explorations, bewley2018learning}. Using synthetic data allows neural networks  to be trained on edge cases in sufficient quantity such that the ML system won't develop undesirable biases. However, such simulations are often \enquote{too easy} for ML systems to understand, and fail in real world deployments~\cite{tan2018simtoreal}. 

To address this issue, we propose the use of machine learning to enhance the outputs of simple simulations, making them perform similarly to much more sophisticated simulators. In our current work with satellite anomaly detection for NASA and NOAA this technique is being developed to create realistic simulations for anomaly detection and classification, but we believe that it is broadly applicable to other agencies. Briefly, the approach is as follows:

\begin{enumerate}
	\item A simulator is constructed that approximately mimics the behavior of the target vehicle. This simulation can be quite coarse - for example a square wave can be used for nearly any periodic waveform, such the rotation of a wheel. This model does not have to include all systems on the target vehicle.
	\item Data, either recorded from operational vehicles or from sophisticated, real-time simulators, is gathered in the course of normal operations. This data represents \textit{baseline} behavior
	\item The simple simulator is configured to generate its version of the baseline data, which is also recorded.
	\item A first neural network is trained to \textit{enhance} the simple data to match the general characteristics of the target data. This model learns to map the coarse behavior of the simulator to a correct, but generalized and unrealistically clean behavior. To add supplementary stochastic information to the output of this network, a second neural network is trained to replicate environmental contributions. The output of both neural networks are combined to produce a high-quality, realistic output.
\end{enumerate}

Once trained, the enhanced simulator can infer realistic signals from a simulator that is running in a variety of \enquote{degraded} configurations. For example, shock absorbers can wear out. Air filters can become clogged. Subsystems can be crippled. Families of vehicles that are built on a common framework can be rapidly generated using the same simulator and different training data. This ability to quickly develop new capabilities that can be run rapidly on commodity hardware allows autonomous ML diagnostic systems to be trained effectively and at scale.

\section{\uppercase{Telemetry example}}
\begin{figure}[H]
	\centering
	\fbox{\includegraphics[width=.7\columnwidth]{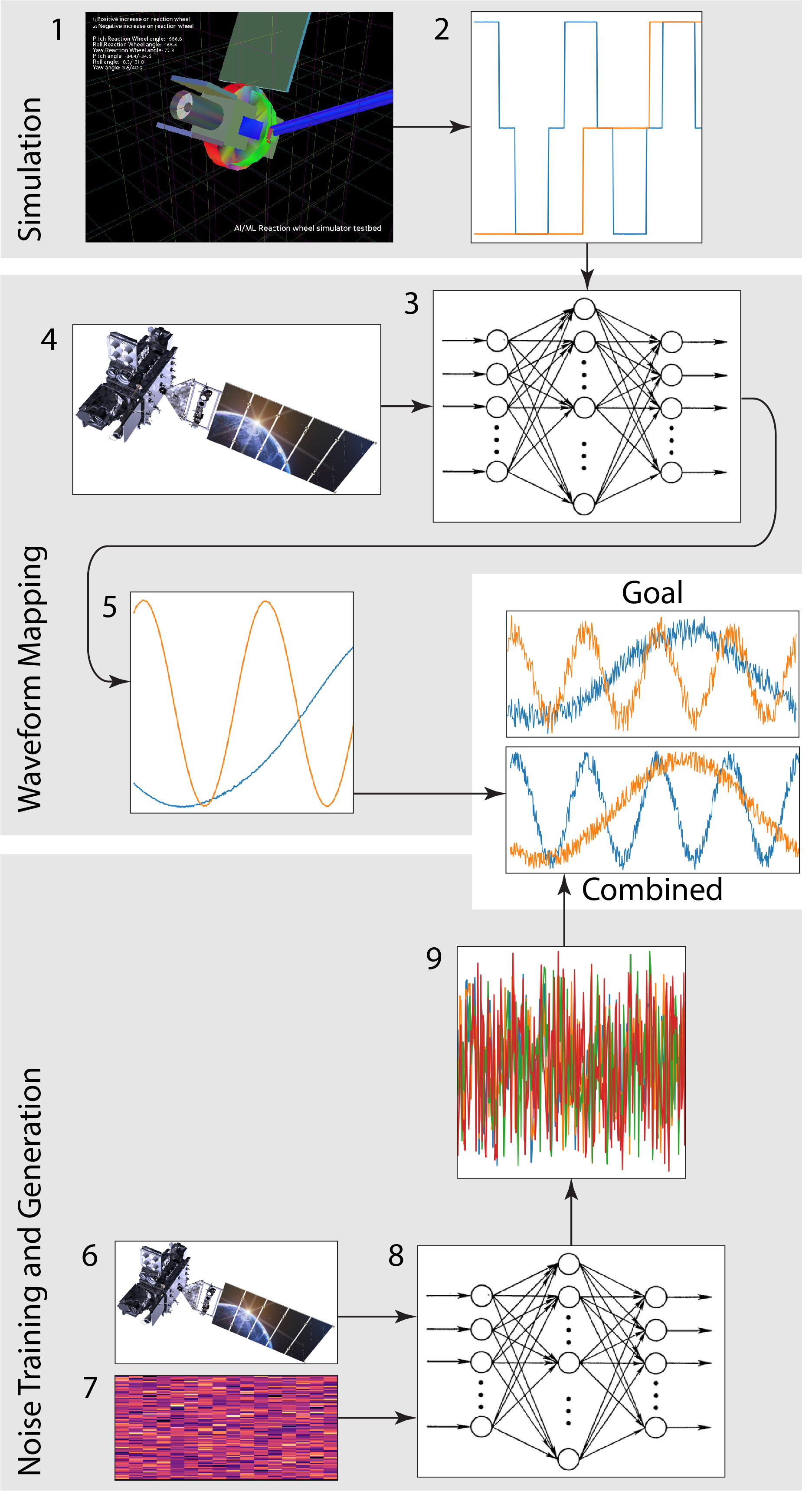}}
	\caption[]{Enhanced simulation pipeline}
	\label{fig:pipeline}
\end{figure}

An overview of the pipeline used to create lightweight, high-fidelity simulations is shown in Figure \ref{fig:pipeline}. For this example, the signals are synthesized sin waves with periods of 8 and 2 minutes. These signals are similar to those generated by rotating satellites (the shorter frequency) in orbit around another object (the longer frequency). To begin, we will generate an example signal and place it in our telemetry storage and retrieval system~\cite{influx2Toolkit}, shown as \enquote{Goal} in figure \ref{fig:pipeline}.  Briefly, the steps involved in the process are:

\begin{enumerate}
	\item Construct, lightweight, high-speed simulations
	\item Generate approximate data
	\item Train a neural network to map low-fidelity to high-fidelity data
	\item High-fidelity simulation or real vehicle data source
	\item Convert simulation to accurate, clean data
	\item Environmental data source
	\item Latent space data source
	\item Train generative adversarial model (GAN) to create realistic environment influence
	\item Generate environmental influences
\end{enumerate}

The elements are then combined to produce the final, high-fidelity output (\enquote{Combined}).

We will apply the same pipeline and techniques to our example data that we would use for actual telemetry.

\begin{comment}
\begin{figure}[H]
	\centering
	\fbox{\includegraphics[width=\columnwidth]{figures/influxSmall}}
	\caption[]{Example telemetry signals in InfluxDB2 database}
	\label{fig:influxsmall}
\end{figure}
\end{comment}

\subsection{Simulation}
%A simple simulation is developed that can approximate the recorded data. The simulation's output is used to feed the input of a neural network which is in turn trained against matching data recorded from the target or generated by more sophisticated simulators if available. 

For this development effort, we had access to highly sophisticated simulators for the NOAA GOES satellites and years of data. These simulators in many cases include the same software and often flight hardware. They are excellent for evaluating a particular set of options given a scenario and are extremely limited with respect to how much faster than real time they can operate. 

\begin{figure}[H]
	\centering
	\fbox{\includegraphics[width=.5\columnwidth]{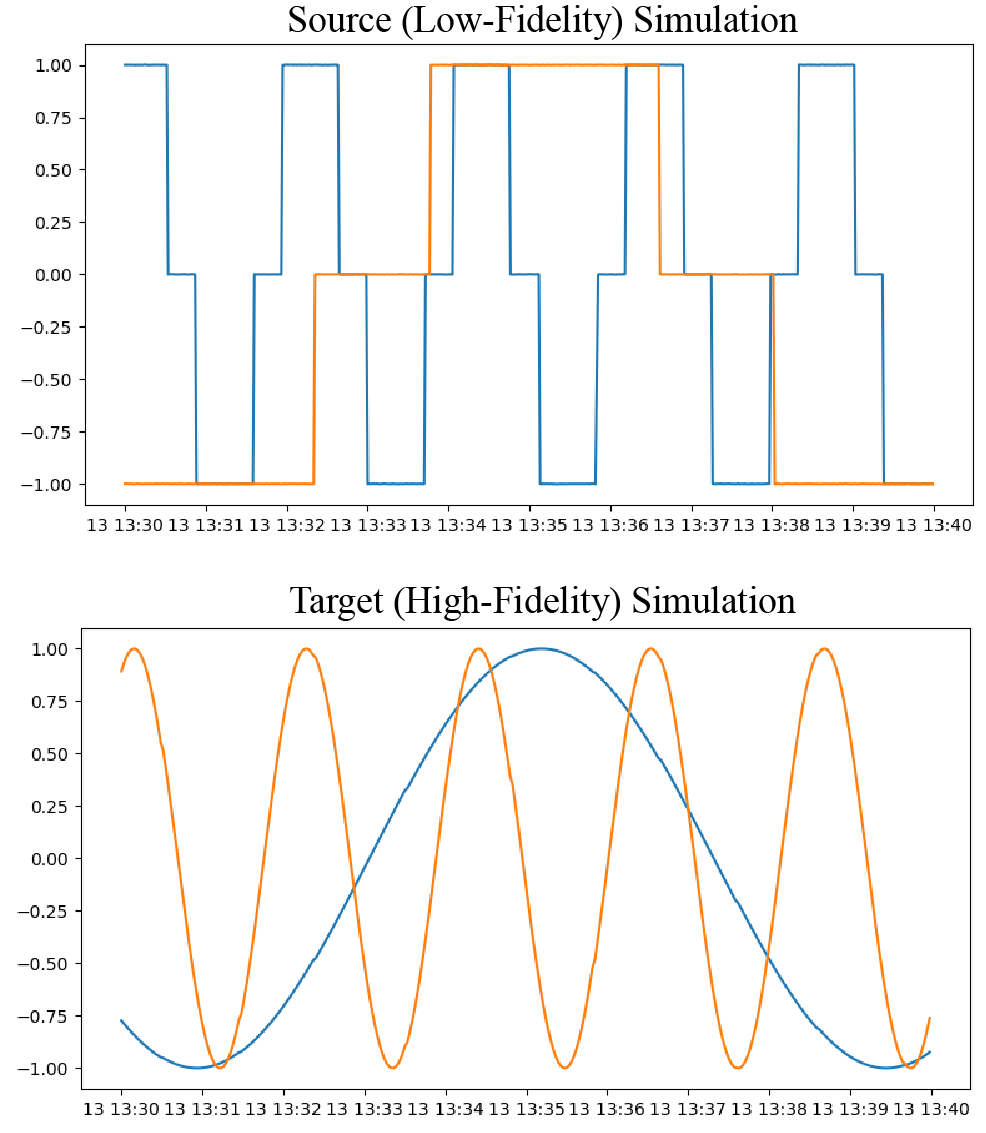}}
	\caption[]{Training Source and Target Time Series}
	\label{fig:source_target}
\end{figure}

ASRC Federal is in the process of developing simple software simulators that can be run in large numbers in the Cloud and much faster than real time. However, a side effect of fast simulators is lower fidelity. So instead of the waveform shown as \enquote{Detail} in Figure \ref{fig:pipeline} that would take 5-10 minutes to produce on a high-fidelity simulator, these simulators can generate the highly quantized data shown in Figure@\ref{fig:source_target} (\enquote{Source}) in a few seconds. 

\subsection{Waveform mapping}
To transform the low fidelity output of the simulator into high-fidelity waveforms while maintaining the computational speed and memory footprint that simple simulations provide requires the training of two neural networks: The first is trained to map the output of the simple simulators to the output of the high-fidelity simulators or recorded activity from the actual satellite. All networks were implemented in Tensorflow version 2.1.0.

\begin{figure}[H]
	\centering
	\fbox{\includegraphics[width=.5\columnwidth]{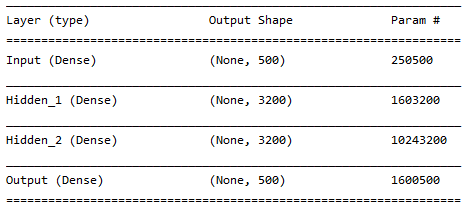}}
	\caption[]{Enhancing MLP Neural Network}
	\label{fig:enhance_nn}
\end{figure}

For this example, we developed a wide, shallow Multi-Layer Perceptron (MLP) network. The structure of the network is shown in Figure \ref{fig:enhance_nn}. It consists of four MLP layers (referred to in Tensorflow as \enquote{Dense}), with two inner, hidden layers (\enquote{Hidden\_1} and \enquote{Hidden\_2}) that then feed to an output layer. The size and number of dimensions are shown in the \enquote{Output Shape} column. Here we can see that these are one dimensional, with the number of neurons indicated by the second value in the tuple. These layers can be fed a variable number of input vectors, as specified by the first, \enquote{None} tag. The input and output layers are the size of the time series. The inner layers are wider, at 3,200 neurons. Wider networks are better at matching functions of this type~\cite{zagoruyko2016wide}. The extra depth is required to match the multiple waveforms that the network has to learn. The last column, \enquote{Param \#}, represents the total number of weights that the network will manipulate during the training sequence. 

\begin{figure}[H]
	\centering
	\fbox{\includegraphics[width=.5\columnwidth]{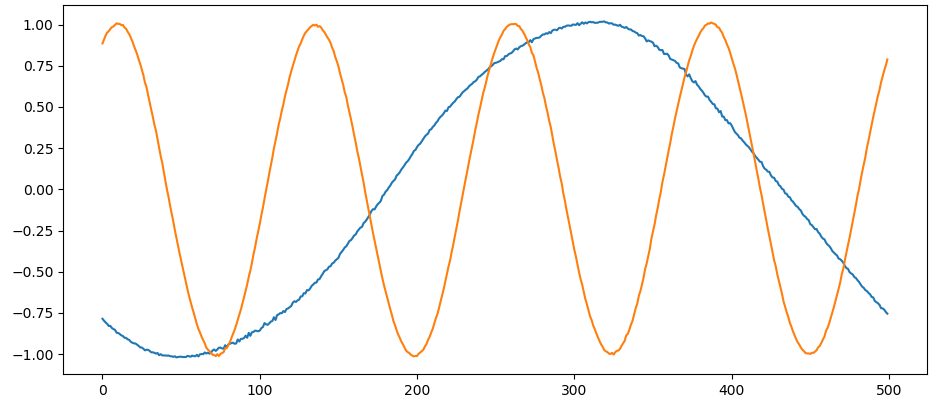}}
	\caption[]{Enhanced Simulation}
	\label{fig:enhanced_sim}
\end{figure}

The model is trained by matching a large number of \enquote{source} time series such as those in Figure \ref{fig:source_target}, with a corresponding set of high-fidelity \enquote{target}time series whose beginning and end are offset by a random amount so that all sample sizes are the same. After training the model for 40 epochs with a batch size of 15, we were able to produce the enhanced waveforms shown in Figure \ref{fig:enhanced_sim}. These waveforms are produced by taking a specific time series of simulation data as a vector (Figure~\ref{fig:source_target} \enquote{Source}) and mapping the input to an enhanced output vector (Figure~\ref{fig:source_target} \enquote{Target}) of the same time. Timing for this output vector can be taken from the corresponding input vector element. Once trained, an input vector is multiplied by these weights to produce the enhanced values shown in figure \ref{fig:enhanced_sim}.

It is important to note that once the model is trained, that the inference that transforms the highly quantized simulation to the smooth, enhanced simulation is extremely fast, particularly when using hardware acceleration. This use of Neural Networks is what allows us to get high-fidelity results out of low-fidelity simulators without substantial speed or memory penalties.

However, we are now at the point that most high-fidelity simulation-based training systems encounter. The signal is too clean. An anomaly detection system trained on signals like these may not be able to discriminate between \enquote{normal} levels of noise and a genuine anomaly. It needs to be processed further to resemble the original waveforms in Figure~\ref{fig:source_target}.

\subsection{Noise Training and Generation}

In this approach, noise is trained independently using Generative Adversarial Neural Networks (GANs)

\begin{figure}[H]
	\centering
	\begin{minipage}{.333\columnwidth}
		\centering
		\fbox{\includegraphics[height = 6em]{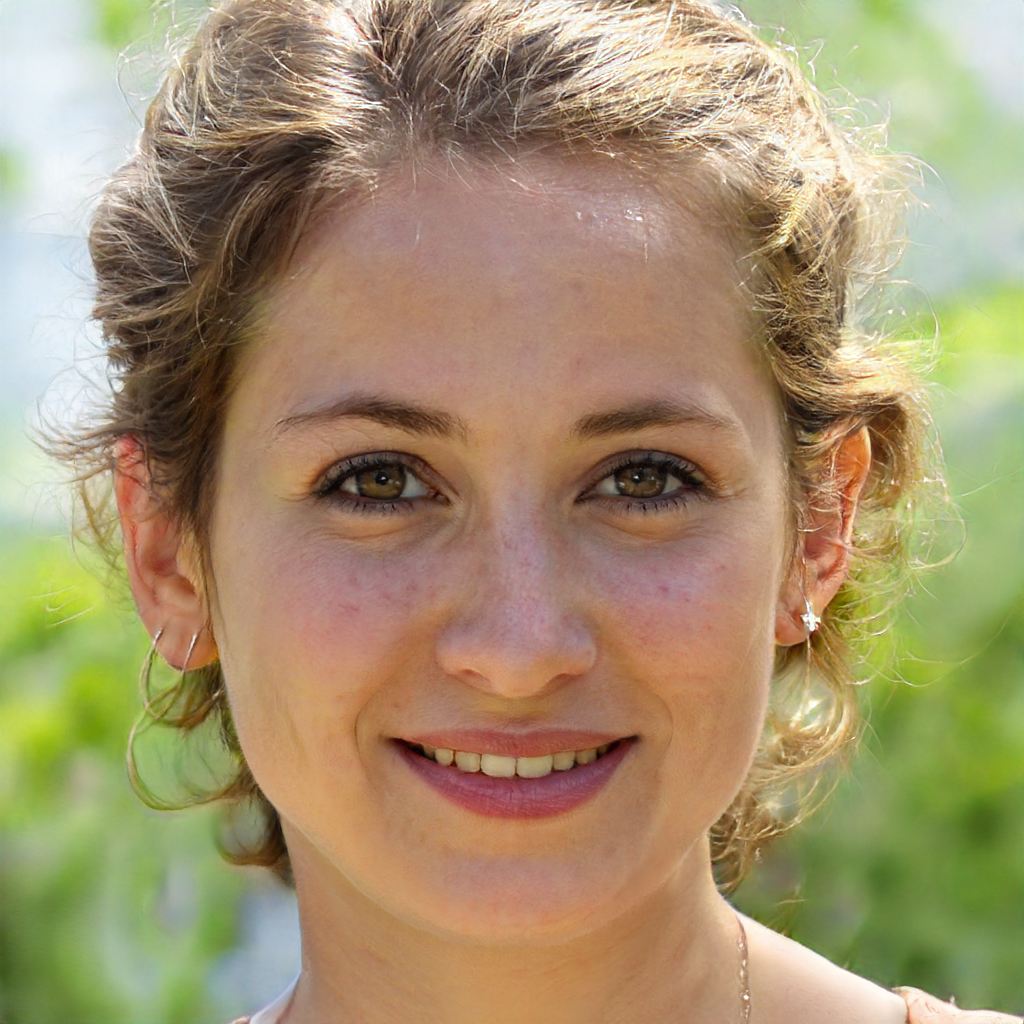}}
		\caption{\label{fig:GANface1}}
	\end{minipage}%
	\begin{minipage}{.333\columnwidth}
		\centering
		\fbox{\includegraphics[height = 6em]{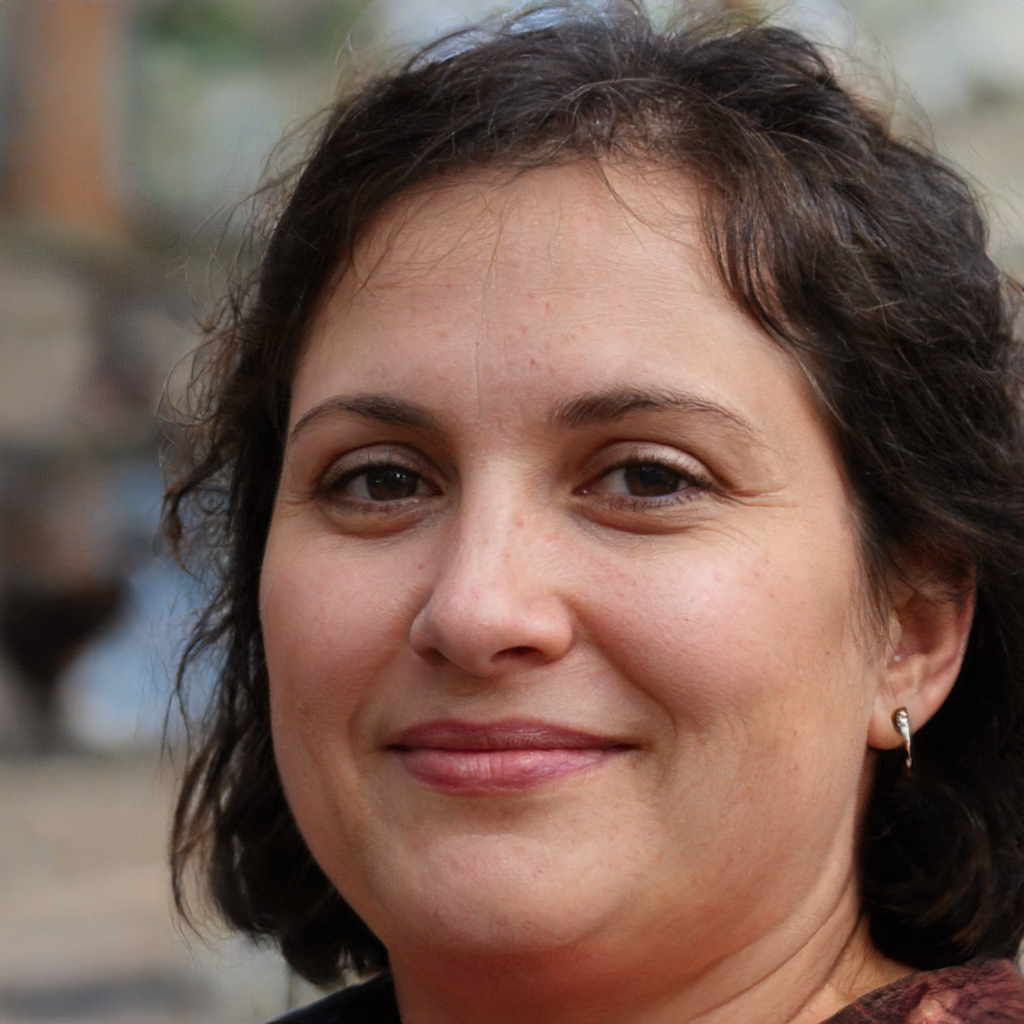}}
		\caption{\label{fig:GANface2}}
	\end{minipage}%
	\begin{minipage}{.333\columnwidth}
		\centering
		\fbox{\includegraphics[height = 6em]{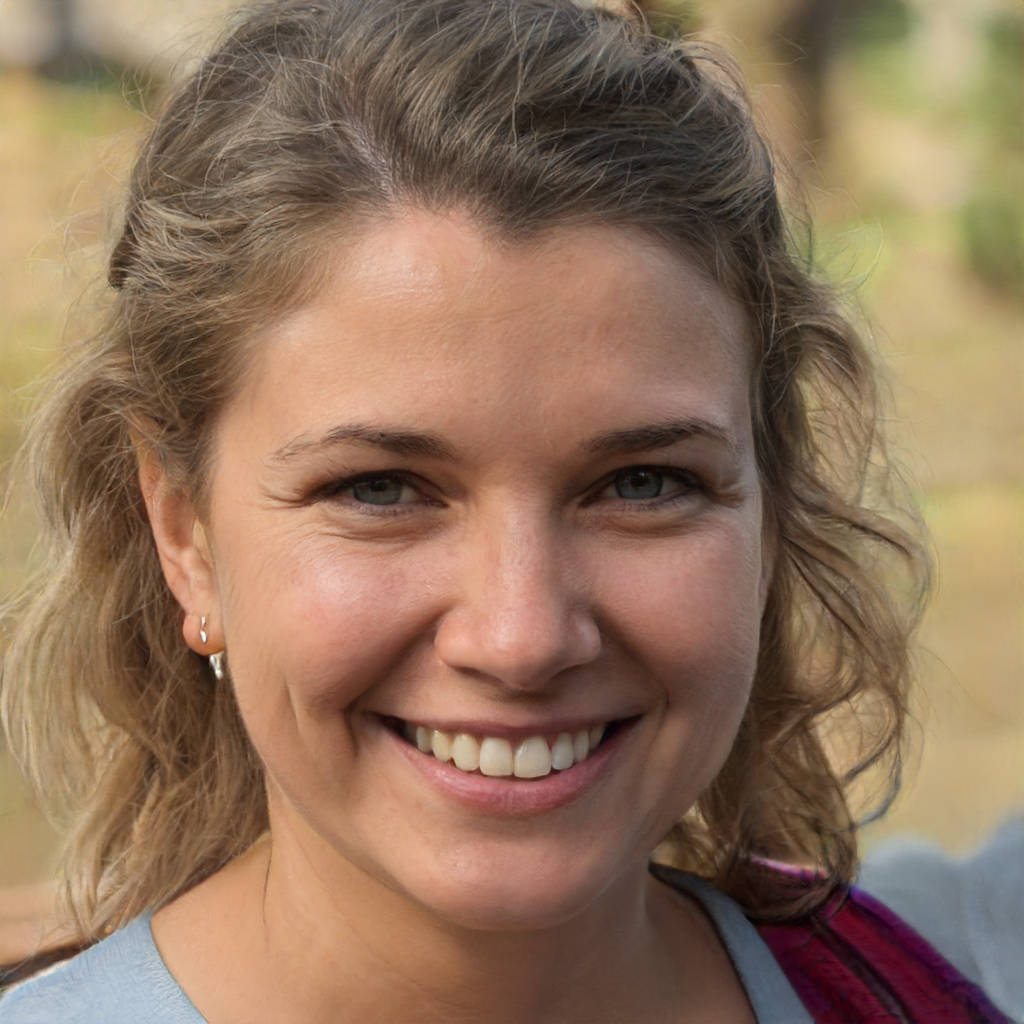}}
		\caption{\label{fig:GANface3}}
	\end{minipage}%
\end{figure}

Generative adversarial learning is a technique where a \textit{generative network} builds synthetic items (such as images) while the \textit{discriminative network} attempts to distinguish the synthetic items from real ones take from a training set or \textit{distribution}~\cite{wang2019deep}. Typically, the generative network learns to map from a randomly generated latent space to the distribution of interest (such as pictures of faces), while the discriminative network tries to detect the synthetic items. The generative network's training objective is to increase the error rate of the discriminative network by \enquote{fooling} the discriminator network through producing synthetic items that the discriminator thinks are real. This technique is quite capable of producing photorealistic results. The faces seen in Figures \ref{fig:GANface1}, \ref{fig:GANface2}, and \ref{fig:GANface3}, are \textit{completely synthetic}, and were generated using the online StyleGAN2 generator \url{https://thispersondoesnotexist.com}~\cite{karras2019style}.

\begin{figure}[H]
	\centering
	\fbox{\includegraphics[width=.5\columnwidth]{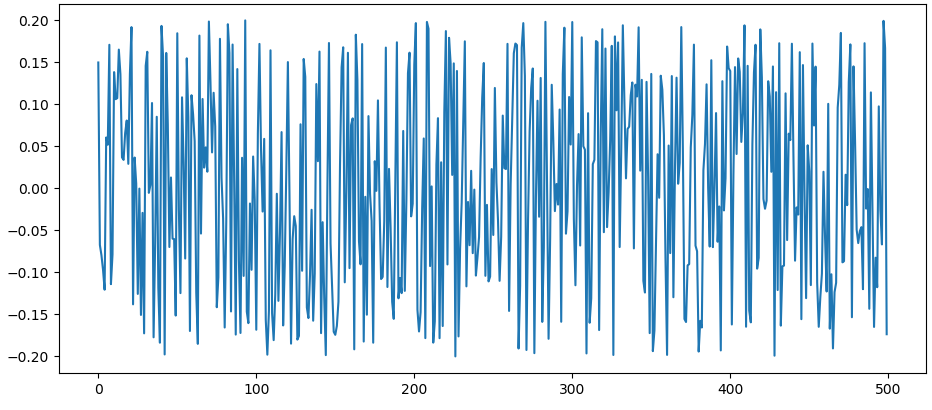}}
	\caption[]{Extracted Source Noise}
	\label{fig:source_noise}
\end{figure}

Noise and other stochastic environmental effects of real telemetry are extracted using a moving average filter~\cite{pandas_rolling_mean_2020}. This average is subtracted from the original signal, leaving the noise that needs to be simulated (Figure \ref{fig:source_noise}).

\begin{figure}[H]
	\centering
	\fbox{\includegraphics[width=.5\columnwidth]{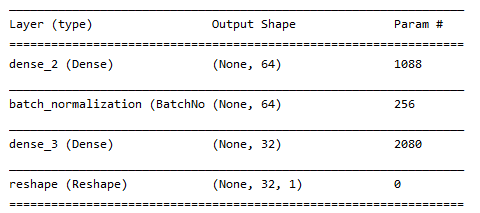}}
	\caption[]{Generator construction}
	\label{fig:generator}
\end{figure}

Our GAN for satellite telemetry is created by connecting the discriminator and generator networks together. The generator was built as shown in Figure \ref{fig:generator}. The 16-element latent vector is densely connected to a layer of 64 neurons. These are then normalized and spread to a layer of 500 neurons, the number of samples in the time series. This layer is reshaped to be compatible as the input layer of the discriminator.

The discriminator is provided with two sources that it must distinguish between:
\begin{enumerate}
	\item A large number of real time series from the database whose beginning and end are offset by a random amount so that all sample sizes are the same.
	\item A matching number of outputs from the generator, using the same format as the real input.
\end{enumerate}

\begin{figure}[H]
	\centering
	\fbox{\includegraphics[width=.5\columnwidth]{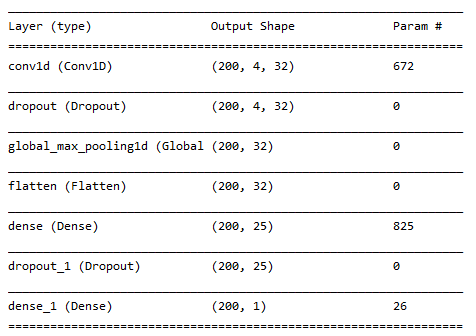}}
	\caption[]{Discriminator construction}
	\label{fig:discriminator}
\end{figure}

The discriminator (Figure \ref{fig:discriminator}) consists of one convolutional layer that merges input from 4 rows of inputs using a window 20 neurons wide with a stride of 4 neurons per step. This layer pools each convolution based on the maximum value of the sample. This configuration aids in finding patterns in signals. The layers are progressively narrowed to the final single neuron who's value determines if the input is considered real or fake. We use a \textit{binary cross-entropy} loss function, which compares the summed errors across all classifications~\cite{rubinstein2013cross} and the \textit{Adam} adaptive optimizer~\cite{kingma2014adam}. This value is then compared against the passed in values from the real and generated inputs to provide the information needed for training.

The real and generated data are tagged for discriminator training and evaluation. Real data has a tag of 1.0, while generated data has a tag of 0.0.

\begin{figure}[H]
	\centering
	\fbox{\includegraphics[width=.75\columnwidth]{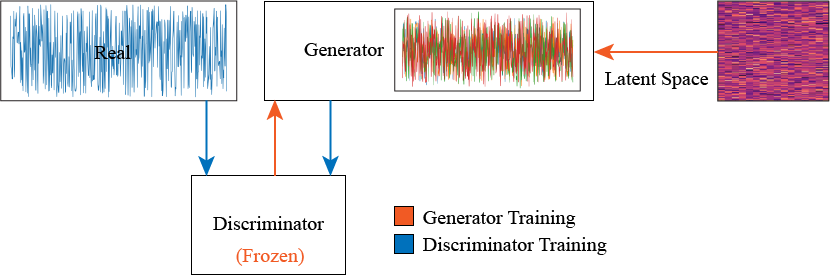}}
	\caption[]{GAN Training}
	\label{fig:gan_training}
\end{figure}

Training is divided into the training of the discriminator, and the training of the generator, as shown in Figure \ref{fig:gan_training}. In the first section, the discriminator as a standalone model is fed with equal amounts of data from the real data set and the generated data set. It then trains on the entire batch (200 rows) of real and generated data. After this pass, the discriminator's weights are frozen, and the generator is trained as part of the entire GAN model. This allows the generator to be trained on the backpropagating error from the discriminator. To have the generator converge on realistic values, the tags for this pass are reversed, and the discriminator is \enquote{told} that the generated values are real. If it determines that they are \textit{false}, then a distance is calculated that would adjust the weights towards the correct answer. Since the discriminator is frozen, the weights are only adjusted on the generator. 

\begin{figure}[H]
	\centering
	\fbox{\includegraphics[width=.75\columnwidth]{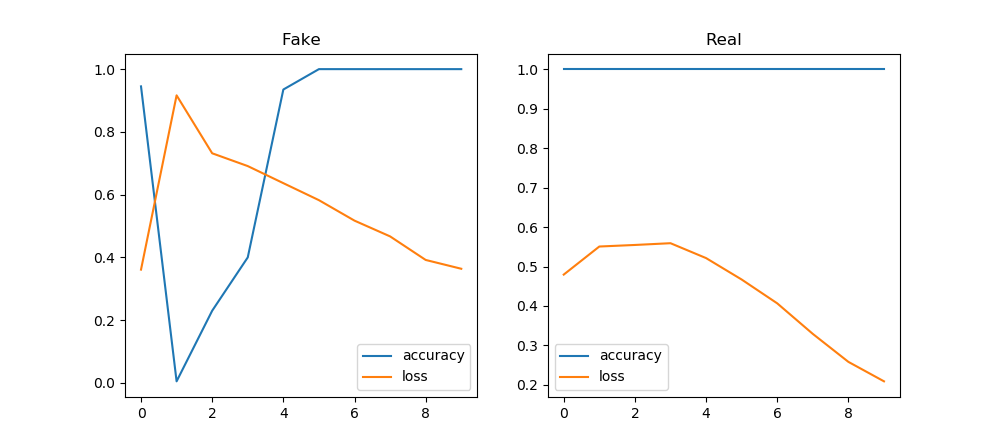}}
	\caption[]{Accuracy/Loss for Real and Fake Classifications}
	\label{fig:acc_loss}
\end{figure}

To match the qualities of this noise, our model needed around 1,000 iterations. During this process. the \textit{accuracy} -- how many of the real and fake samples were correctly classified, and the \textit{loss} -- the normalized error across all classifications were sampled at 100-step intervals across the 1,000 iterations and are shown in Figure \ref{fig:acc_loss}. It is important to remember when looking at this chart that the generator and the detector are engaged in an \textit{adversarial} process, where the generator constantly tries to improve its ability to fool the detector, and the detector constantly tries to improve its ability to identify these forgeries. As we can see in the figure, the discriminator improves slightly faster than the generator, which is the goal of a GAN. If the two elements are too imbalanced, the system cannot learn effectively.

After 1,000 iterations, the generated noise is sufficiently similar to the actual noise. The output of the simulator enhancing neural network can then be summed with the noise-generating neural network to produce the final signal, shown in Figure \ref{fig:combined}, below:

\begin{figure}[H]
	\centering
	\fbox{\includegraphics[width=.6\columnwidth]{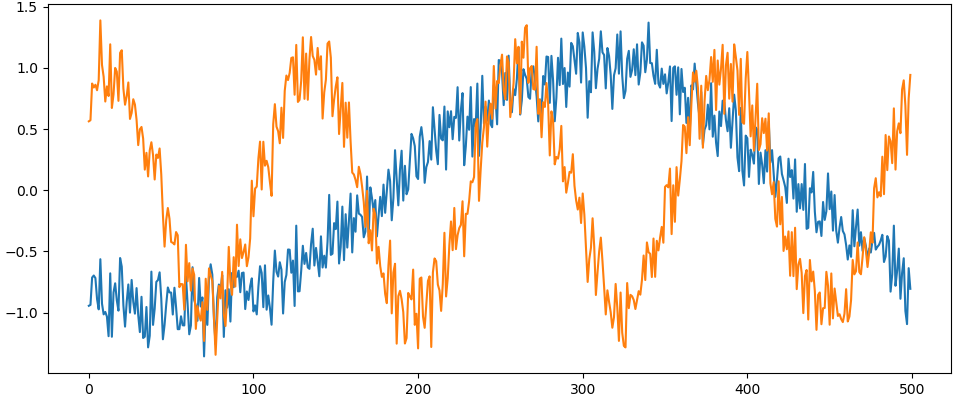}}
	\caption[]{Simulator output, enhanced, with noise added}
	\label{fig:combined}
\end{figure}

\section{\uppercase{Discussion}}
The majority of this particular research occurred during the peak of the COVID-19 NASA/NOAA response, and we were unable to access the high fidelity simulators we were planning to use. As such the simulations described in this paper are based on simplified approximations. However, based on our experience modeling other satellite telemetry with small, rapidly-trained MLP networks~\cite{li2017machine} leads us to believe that the results here can be applied to actual telemetry data when it becomes available.

An issue that needs to be examined in more detail is the ability for the enhancing network to adapt low-fidelity signals to cases not covered by real telemetry or high-fidelity simulators. Because high-fidelity simulators are rare, they will only be used to explore likely problem spaces, or respond to situations that occur on the actual vehicle. Unusual edge cases that show up quickly where only the low fidelity simulators are capable of responding will have to be researched more deeply to see if the data produced by the enhancing neural network is sufficiently valid.

An example of such an unusual scenario occurred in 2009, when US Airways flight 1549 struck a flock of geese shortly after takeoff resulting in a loss of power in both engines. With only a short period of time to evaluate potential options, the flight crew decided to ditch the plane in the Hudson river, saving all passengers~\cite{wiki:Flight_1549}.

What would have happened if the crew had been less experienced? Could there be a way to evaluate options for these types of cases where time is critical and experience limited or nonexistent? If enhanced simulators can be built to be small and fast enough to run at many times normal speed and in parallel, it may be possible to automate a response to a Mayday request by starting an always-available cluster of simulations to evaluate potential best options given edge-case degraded modes. In essence, multiple reinforcement learning simulations are set up with the objective function being in the case of Flight 1549, a safe landing. 

%TODO: tie baxck to ground vehicles

Such simulations need not be limited to satellites or civil aviation. Simulation and prediction of degraded ground vehicle behavior ranges from situations as specific as overheating train axles~\cite{yang2019research} to predicting traffic~\cite{jiber2018traffic}. Combat often involves ground vehicles operating individually or in groups in degraded modes that cannot anticipated. An approach to adapting quickly to these unanticipated situations using scalable high-fidelity simulation may make for a more adaptive combat capability that is able to adjust to changing conditions faster than the Adversary.

\section{\uppercase{Conclusions}}
All machine learning depends on large volumes of data. Creating a pipeline for providing synthetic data on demand is a market that is currently worth tens of millions of \$US annually that is likely to only  increase over time. ASRC is developing systems to provide synthetic data at scale. Synthetic data allows organizations to be independent of data sources with potential limitations and foreign complications. 

Machine learning models are useless without data, and diverse data can make the same model applicable in diverse contexts. Even if progress ceased in the development of more sophisticated models, machine learning could be effectively applied to new domains simply by training current state-of-the-art models with new, well understood and balanced datasets.

Simulation as a way of creating usable assets is currently being done in an ad-hoc basis in the AI/ML community. Particularly for the government user, it is often the only secure way to generate the amounts of data needed for the effective training of unusual models, such as satellites. In this paper, we have shown that it may be feasible to produce large amounts of simulated data that can in turn be used to train machine leaning systems to recognize and adapt to rare and unlikely situations. Future work will focus on increasing the range, scale, and sophistication of these types of simulations.

% Bibliography

%\newpage
%\input{addendum}
%\newpage

\end{document}